%% file: CharacterGAN.tex
\documentclass[10pt,twocolumn,letterpaper]{article}

\usepackage{wacv}

\usepackage{times}
\usepackage{epsfig}
\usepackage{graphicx}
\usepackage{amsmath}
\usepackage{amssymb}
\usepackage{booktabs}
\usepackage{caption}
\usepackage{verbatim}
\usepackage[final]{animate}
\usepackage{xspace}
\usepackage{enumitem}
\usepackage{microtype}

\wacvfinalcopy 

\ifwacvfinal
\usepackage[pagebackref=true,breaklinks=true,colorlinks,bookmarks=false]{hyperref}
\else
\usepackage[pagebackref=true,breaklinks=true,colorlinks,bookmarks=false]{hyperref}
\fi

\newcommand{\myparagraph}[1]{\vspace{.2cm} \noindent \textbf{#1} \quad}

\DeclareRobustCommand{\animation}{%
   \includegraphics[height=\fontcharht\font`\B]{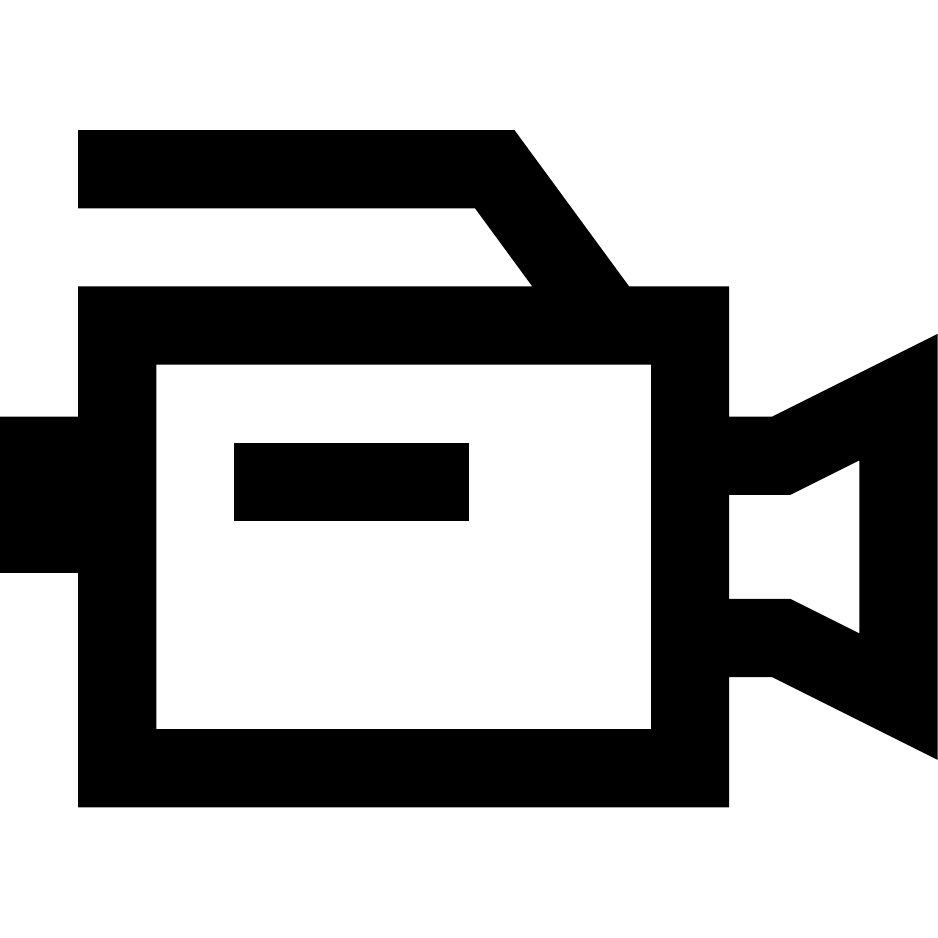}\hspace{-.2em}
}

\begin{document}

\title{CharacterGAN: Few-Shot Keypoint Character Animation and Reposing}

\author{Tobias Hinz$^{1,2}$\\
\and
\hspace{-2em}
Matthew Fisher$^2$\\
\hspace{-2em}
$^1$University of Hamburg\\
\and
\hspace{-1em}
Oliver Wang$^2$\\
\and
\hspace{-1em}
Eli Shechtman$^2$\\
\hspace{-1em}
$^2$Adobe Research\\
\and
Stefan Wermter$^1$\\
}

\twocolumn[{%
\renewcommand\twocolumn[1][]{#1}%
\maketitle
  \centering
  \input{fig_teaser}
  \captionof{figure}{We train a generative model in a low-data setting (8 to 15 training samples) to repose and animate characters based on keypoint positions. The first row shows \textbf{video} results (indicated throughout the paper by \animation), of our method driven by linearly interpolating input keypoints. Please view with Adobe Acrobat to see animations. The second and third rows show interpolated frames generated by our method between a \emph{single} start and end frame (left and right columns).}
  \vspace{1em}
  \label{fig:teaser}
}]

\maketitle
\thispagestyle{empty}

\begin{abstract}
\vspace{-1em}
We introduce CharacterGAN, a generative model that can be trained on only a few samples (8 -- 15) of a given character.
Our model generates novel poses based on keypoint locations, which can be modified in real time while providing interactive feedback, allowing for intuitive reposing and animation.
Since we only have very limited training samples, one of the key challenges lies in how to address (dis)occlusions, e.g.\ when a hand moves behind or in front of a body.
To address this, we introduce a novel \textit{layering approach} which explicitly splits the input keypoints into different \textit{layers} which are processed independently.
These layers represent different parts of the character and provide a strong implicit bias that helps to obtain realistic results even with strong (dis)occlusions.
To combine the features of individual layers we use an adaptive scaling approach conditioned on all keypoints.
Finally, we introduce a mask connectivity constraint to reduce distortion artifacts that occur with extreme out-of-distribution poses at test time.
We show that our approach outperforms recent baselines and creates realistic animations for diverse characters.
We also show that our model can handle discrete state changes, for example a profile facing left or right, that the different layers do indeed learn features specific for the respective keypoints in those layers, and that our model scales to larger datasets when more data is available.
Code is available at \url{https://github.com/tohinz/CharacterGAN}.
\end{abstract}

\vspace{-1em}
\section{Introduction}
Modifying and animating 2D artistic characters is a task that requires experts to manually create many instances of the same character in different poses, which is a time-consuming and expensive process.
As a result, often only very few examples of a specific character are available, as the cost of acquisition of additional examples is high.
Using video frame interpolation methods to create new poses can reduce the overhead, but these methods do not leverage character-specific priors and so can only be used for small motions. 
Similarly, warping input frames directly with 2D handles cannot account for disocclusions or appearance changes between poses. 
We believe that control over 2D animated characters is a challenging open problem, due to the sparse example set and challenging topology changes.
In this work we have two main goals: (1) to generate high quality frames of an animated character based on a small number of examples, and (2) to generate these images based on a sparse set of keypoints that can be easily modified in real time.

To address both issues, we propose to train a conditional Generative Adversarial Network \cite{goodfellow2014generative} (GAN) that allows us to create new images of a character based on a set of given keypoint locations.
We show that, with the right form of implicit biases, such a model can be trained in a few-shot setting (i.e., 10s of training images).
In character animation, each training image has to be created manually, making it expensive to acquire the number of images required for training most conditional GAN models~\cite{isola2017image}, and unlike recent work on single image generative models~\cite{shaham2019singan}, we desire precise control over the generated image.
Our approach makes it possible for the broad community to animate and repose any 2D character without the need for specialized skills.

Our method consists of a GAN that is trained on 8 -- 15 images of a given character and its associated keypoints.
One of the key challenges is that it is difficult for the model to learn which parts of a character should be occluded by other parts.
E.g.\ when the hand of a humanoid character moves in front of the torso, either the hand should be visible at all times (if it is in the foreground) or only the torso should be visible if the hand moves ``behind'' the torso.
To learn this ordering in a purely data-driven approach we need many images which we do not have in our setting.
Instead, we propose to use user-specified \textit{layers} for our keypoints, i.e.\ each keypoint lies in a given layer and we can introduce an ordering over those layers.
For example, a humanoid character could be described by three layers, consisting of 1) the arm and leg in the ``back'' (i.e.\ occluded by the torso and other arm and leg), 2) the torso and head which are occluded by one arm and one leg and occlude one arm and one leg, and 3) the other arm and leg which occlude every other part of the character.
Our model processes each of these layers independently, i.e.\ generates features for each layer without knowing the location of the keypoints in the other layers.
We then use an adaptive scaling approach, conditioned on all keypoints, to spatially scale the features of each layer, before concatenating and using them to generate the final image.

Additionally, we can train our generator to predict the mask for a generated character which we found to be a robust way to identify and automatically fix unrealistic keypoint layouts at test time.
Our model naturally learns to associate the keypoints with roughly semantically meaningful body parts for each layer, and can handle discrete state points that arise as a function of keypoint locations (e.g., switching between a profile facing left or right).
We show via a number of qualitative and quantitative experiments that our resulting model allows for real-time reposing and animation of diverse characters and that our layering approach outperforms the more traditional conditioning approach of using all keypoints at once. 
Since we assume no prior knowledge about the modeled character our model can be applied to any shape and does not require additional data such as a 3D model or a character mesh.
This allows our model to be applied in domains for which only limited data is available (e.g.\ artistic drawings or sprite sheets) without the need for additional manual input besides the keypoint labels.
In summary, we introduce the following main contributions:
\begin{itemize}[noitemsep]
    \item We show that it is possible to train a GAN on only few images (8 -- 15) of a given character to allow for few-shot character reposing and animation. By only conditioning the training on keypoints (instead of e.g.\ semantic maps) the trained model allows for character reposing in real-time without expert knowledge.
    \item By using a layered approach that explicitly encodes the ordering of different keypoints our model is able to model occlusions with only very limited training data.
    \item We introduce a mask connectivity constraint, where a jointly predicted mask can be used at test time to automatically fix keypoint layouts for which the model produces unrealistic outputs.
\end{itemize}

\section{Related Work}
\vspace{-0.5em}
Conditional GANs make it possible to control the output of the generator to varying degrees and have also been shown to help with the training process.
The conditioning input can come in several forms, such as class conditioning \cite{mirza2014conditional}, semantic maps \cite{isola2017image}, keypoints \cite{reed2016learning}, or bounding boxes \cite{hinz2019generating,hinz2020semantic}.
However, most conditional GANs are trained with large datasets and are applied to broader domains, whereas we are interested in animating a \emph{specific} character.
In the following, we first focus on approaches in how to leverage small amounts of data to train GANs and conclude with a section about how the chosen conditioning method directly affects how the model can be used at test time.

\begin{figure*}
    \centering
    \includegraphics[width=0.9\textwidth]{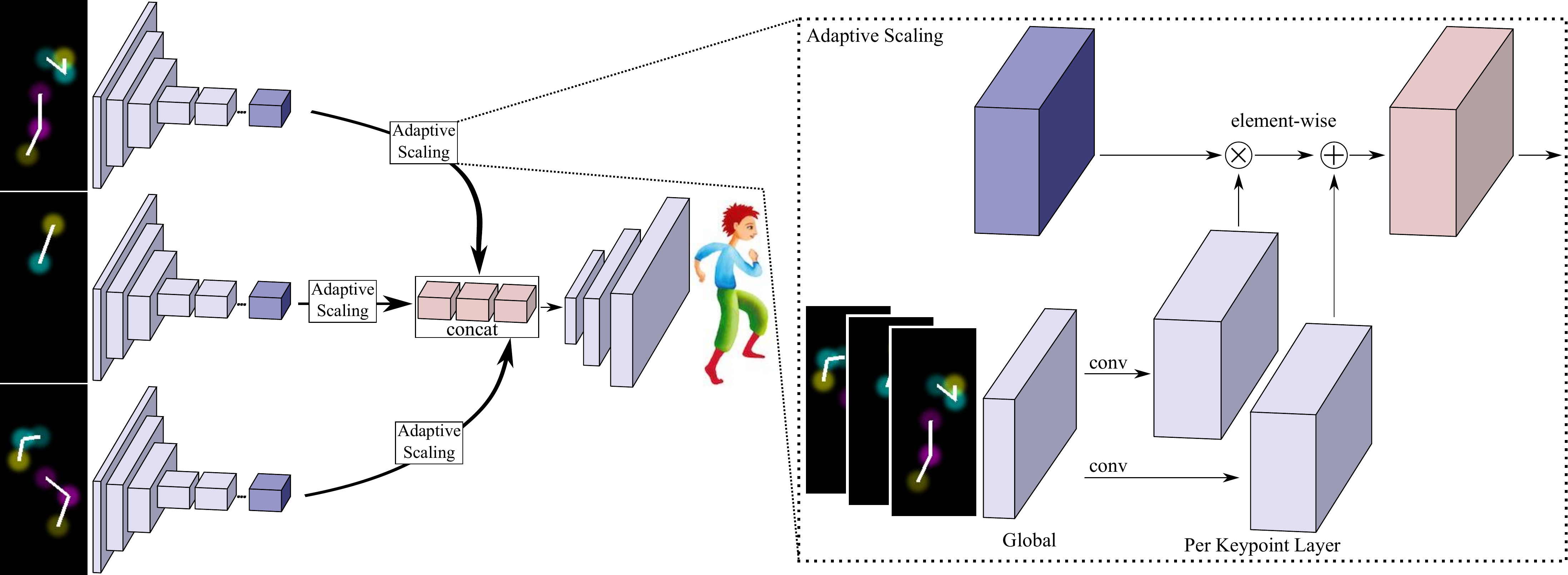}
    \vspace{-0.5em}
    \caption{Our model processes keypoints which are split into individual layers. The resulting features for each keypoint layer are then scaled, concatenated, and used to generate the final image.}
    \label{fig:model_overview}
    \vspace{-1em}
\end{figure*}

\myparagraph{Few-Shot Learning with GANs}
One promising approach to few-shot learning with GANs is to fine-tune GANs that are pre-trained on large datasets.
This can be achieved by fine-tuning a given model \cite{wang2018transferring}, by only training parts of the pre-trained model \cite{noguchi2019image,robb2020few,mo2020freeze,li2020few}, or transfering knowledge between different - but related - domains \cite{wang2020minegan,zhao2020leveraging}.
However, these methods still rely on a model that is pre-trained on a large dataset in a very similar domain.
In contrast to this, we do not fine-tune a pre-trained model but instead train our model from scratch on limited available data.

Recent approaches show that applying data augmentation techniques during training of GANs directly is useful, especially when the dataset is small \cite{tran2020towards,karras2020training,zhao2020differentiable,zhao2020image}.
One of the main insights of these approaches is that it is not sufficient to only augment real images since this leads the discriminator to learn that these augmented images are part of the real data.
We also make heavy use of data augmentation in our training process but use much less data (8 -- 15 images) than the approaches mentioned here (usually 100+ images).

It is also possible to learn useful features from a single image \cite{zontak2011internal,asano2020critical} and tasks such as texture synthesis \cite{jetchev2016texture,li2016precomputed,bergmann2017learning,zhou2018non}, image retargeting \cite{shocher2019ingan}, inpainting and segmentation \cite{ulyanov2018deep,gandelsman2019double}, and unconditional image synthesis \cite{shaham2019singan,hinz2020improved} are possible with only a single image.
Other approaches train GANs for image-to-image translation with only one pair of matching images \cite{lin2020tuigan,benaim2020structural,park2020contrastive,vinker2020deep}.
However, a model trained on just a single image can only generate limited variations of the training input.
Therefore, we propose to increase the available information by slightly increasing the training set's size, which allows the model to generate more variations of a learned object.

\myparagraph{Editability}
An important characteristic of models for character reposing and animation is how the character is controlled.
Existing approaches use driving videos \cite{siarohin2019first}, learn a distribution over poses for inbetweening \cite{poursaeed2020neural}, or map puppets to skeletons \cite{dvoroznak2018toonsynth}.
However, it is difficult to achieve a desired pose with these approaches \cite{siarohin2019first,poursaeed2020neural} or requires expert knowledge to obtain the training data \cite{dvoroznak2018toonsynth}.
Another approach is to condition the model on a semantic map \cite{chan2019everybody,vinker2020deep} which can be changed at test time.
However, modifying semantic maps or edge maps is not something that can be done on-the-fly, but instead takes time and skill.
Finally, one could start with the character in a given pose and warp or stretch it until it reaches the desired pose \cite{liu2014skinning} which might lead to unrealistic results if the end pose is too different from the starting pose.
Given the previous limitations we only condition our model on high-level keypoints which are provided for a given character, making it easy and fast to generate new poses at test time by moving keypoints.

\section{Methodology}
\vspace{-0.5em}
We focus on character animation and reposing where the images show the same character in different poses from the same viewpoint.
Since we work in low data scenarios, we focus on cases with a small amount of images of the given character.
Furthermore, the individual images can depict discrete appearance variations, e.g. different facial expressions.
We do not make any other prior assumptions about the structure of the character.
Our method requires an input image set, keypoints per image, keypoint connectivity information (essentially a skeleton), and a layer ordering for the keypoints. 
This information can be obtained easily as we only need labels for very few images.
While the keypoints have to be labeled manually, all other information is image independent and, thus, only needs to be defined once.

We experimented with various ways to input keypoints to the network, such as as individual channels, but found that representing keypoints as RGB Gaussian blobs performed the best.
This also means that the input condition always has the same input dimensionality, independent of the number of keypoints, which is helpful as our model uses the same parameters to process these individual layers.
Each keypoint is defined by a position $\{x, y\}$, three randomly chosen color values for the three RGB channels, and the values $\sigma$ and $r$, where $\sigma$ represents the falloff of the Gaussian distribution we use for blurring the keypoint, and $r$ represent the radius of the final keypoint, which we define later.
Discrete keypoints are modelled with individual RGB values, and in cases where different keypoints overlap we sum the respective colors.

\subsection{Model}
\vspace{-0.5em}
Our model consists of a generator and a discriminator (see \autoref{fig:model_overview}).
The generator generates the image based on the keypoint locations while the discriminator is trained to distinguish between real and fake image-keypoint pairs.
To address the challenge of having only a limited number of training images we incorporate several problem-specific implicit biases into our generator design and combine this with data augmentation techniques.

\myparagraph{Generator and Discriminator}
We base our architecture on pix2pixHD~\cite{wang2018high}.
However, while the discriminator is similar to the original pix2pixHD discriminator, we add several implicit biases to the generator reflecting our prior knowledge about the problem.
Concretely, we know that the modeled characters are inherently three-dimensional, i.e.\ if some body parts are occluded by others they still exist even though they may not be visible.
To address this, we split our characters into different layers, e.g.\ representing the ``left'' side of the character (e.g.\ left arm and leg), the ``middle'' part of the character (e.g.\ head and torso) and the ``right'' side of the character (e.g.\ right arm and leg).
These layers can be modeled individually and are then be composed to form a final image.
Our generator processes each keypoint layer individually and, thus, learns layer-specific representations.

Intuitively, we could set some features to zero if they are occluded by other features.
For example, if the left hand is ``behind'' the torso for a given image, zeroing out the features for the left hand might make it easier for the generator to generate a realistic image.
Conversely, if the right hand is ``in front of'' the torso, zeroing out the torso features at the location of the right hand might improve the performance.
To address this, we incorporate an adaptive scaling technique \cite{park2019semantic} in which we scale the features of each layer before concatenating them.
For this, we first learn an embedding of the keypoints and their layers (``Global'' in \textit{Adaptive Scaling} of \autoref{fig:model_overview}).
Based on this embedding we then learn scaling parameters for each keypoint layer and use them to scale the features of each each layer.
These scaled layer features are then concatenated and used to generate the final image.

Our discriminator takes the keypoint conditioning $k$ concatenated with an RGB image as input and classifies it as either real or fake.
We use two patch-discriminators \cite{isola2017image}, one of which operates on the full resolution image, while the second operates on the image down-scaled by a factor of two.
We use a feature matching and an adversarial loss during training as defined by the pix2pixHD model \cite{wang2018high}.
The adversarial loss is the standard GAN loss $L_{\text{adv}}$:
\begin{equation}
\begin{aligned}
    \underset{G}{min}\ \underset{D}{max}\  \mathcal{L}_{\text{adv}} = \mathbb{E}_{(k,x)}[\text{log}D(k,x)] + \\
    \mathbb{E}_{(k)}[\text{log}(1-D(k, G(k))]
\end{aligned}
\end{equation}
where $k$ is the keypoint condition and $x$ is the corresponding real image.
The feature matching loss $L_{\text{fm}}$ stabilizes training by forcing the generator to produced realistic features at multiple scales and is defined as:
\begin{equation}
    \underset{G}{min}\ \ \mathcal{L}_{\text{fm}} = \mathbb{E}_{(k,x)}\sum_{i}^{T}\frac{1}{N_i}[\vert\vert D(k,x) - D(k, G(k))\vert\vert_1 ]
\end{equation}
where $T$ is the number of layers in the discriminator and $N_i$ is the number of elements in each layer.
We add a perceptual loss \cite{johnson2016perceptual,zhang2018unreasonable} to further improve the image quality.
We use a VGG net to extract features from real and generated images and compute the perceptual loss $L_{\text{perc}}$ as defined by \cite{johnson2016perceptual}.
Our final loss is the combination of these losses:
\begin{equation}
    \underset{G}{min}\ \underset{D}{max}\  \mathcal{L}_{\text{adv}} + \mathcal{L}_{\text{fm}} + \mathcal{L}_{\text{perc}}.
\end{equation}

\input{tab_baselines}
\input{tab_ablation}

\myparagraph{Data Augmentation}
\label{sec:dataaugment}
We employ both affine transformations and thin-plate-spline augmentation \cite{vinker2020deep}.
We use a mixture of horizontal and vertical translations and horizontal flipping and randomly sample a subset from these augmentation approaches at each training iteration.
Thin-plate-spline (TPS) augmentation was introduced by Vinker et al. \cite{vinker2020deep}.
For this approach, the image is modeled as a grid and each grid point is then shifted by a random distance sampled from a uniform distribution.
After this, a TPS is used to smooth the transformed grid into a more realistic warp.
Using TPS augmentation results in warped images where parts of the image are stretched and elongated, adding further variation to the training data.
All augmentations are applied to the given image and the associated keypoints and skeleton.

\subsection{Editability}
\vspace{-0.5em}
After our model is trained it offers a straightforward way to modify the pose of the character.
Given an image of the character the user can drag keypoints to novel positions and the generator will generate the character in the new pose.
We can also easily switch between different discrete states, and we provide two ways to do this. 
First, it can be handled completely automatic where the discrete state is determined solely by keypoint positions (e.g., for facing left vs right).
Second, we provide the ability to have specific keypoints for individual states, such as smile vs frown, so a user can choose the desired expression at inference time. 
We also allow the user to optionally enable a mask-based connectivity correction. 
In this case, if the user positions keypoints too far from their input distribution, such that they could lead to unrealistic or undesired results, we can automatically modify the keypoint locations to achieve more realistic results.

\myparagraph{Ensuring Mask Connectivity}
If the image background is of uniform color, or we have a segmentation network, we can automatically extract a foreground-background mask.
We can use this mask as additional conditioning information during training, i.e.\ in this case the generator does not only generate the RGB image but also the mask, while the discriminator gets as input an image and its associated keypoints and mask.
At test time, if a keypoint is moved in a way that results in a layout that is too different from the ones seen at train time it can happen that the generator generates either ``disconnected'' body parts (e.g.\ the hand is not connected to the body) or introduces unwanted artifacts.
Since the generator predicts the mask for the generated character we can use connected component analysis \cite{fiorio1996two,wu2005optimizing} to check whether the generated mask is connected.
We found two cases that often lead to a disconnected mask; if the disconnected part contains a keypoint the resulting character will be ripped apart or, alternatively, if the disconnected part does not contain a keypoint the generated image will often contain unwanted artifacts.
In either case, we fix the result by automatically moving nearby keypoints in an iterative procedure until the predicted mask is fully connected again.

Given the last moved keypoint that caused the a disconnected region to appear, we identify the closest keypoint and move this keypoint in the same direction as the keypoint moved by the user by a fraction $\delta$ of the absolute distance.
If $\delta = 1$ the keypoint is moved exactly in the same direction and by the same amount as the original keypoint and if $\delta = 0$ no keypoints are moved automatically at all.
We found that a default of $\delta = 0.5$ achieved good results in a single iteration in most cases, however, this process can be repeated iteratively until the mask is connected (\autoref{fig:fix_kp}).

\input{fig_our_model_examples}

\input{fig_reconstruction}

\section{Experiments}
\vspace{-1em}
\myparagraph{Baselines}
Since this specific problem has not been addressed in the literature there are no standard baselines available to compare against and, therefore, we chose to use three single-image models as our baselines as they are closest to our settings regarding needed data.
SinGAN \cite{shaham2019singan} and ConSinGAN \cite{hinz2020improved} are trained on a single image for tasks such as image generation and harmonization.
Both models are trained in a multi-stage manner where the image resolution increases with each stage.
We adapt both models by additionally conditioning the training at each stage on the image keypoints.
DeepSIM \cite{vinker2020deep} trains a model specifically for image manipulation, where the conditioning input consists either of edge-maps, semantic labels, or a combination of both and the training set consists of a single image and its condition.
We adapt the DeepSIM model to our setting by replacing the edge-map conditioning with keypoints.

\input{fig_animation}

\myparagraph{Data}
We perform experiments on several different characters, including humans and animals.
Our data comes from different sources such as drawings by artists \cite{dvoroznak2018toonsynth} and characters from sprite sheets.
For our characters we have 8 -- 15 images which we manually label with keypoint locations.
All manual labeling can be done without any specific skills and our approach requires only RGB images as input.
In contrast, traditional animation pipelines would likely need additional expert drawings, as well as a more structured character representation (e.g. a mesh).
As a result, our method is less labor-intensive, and is more amenable to unskilled users compared to traditional animation approaches.

We use the same layer definition and number of layers (three) for all our examples, roughly corresponding to ``front'', ``middle'', and ``back''.
Based on our experiments, the keypoint labeling and model training for a new character can be done in well under one hour.
Manually labeling new keypoints takes roughly 30-60 seconds per image (depending on the number of keypoints and used interface) while training the model takes roughly 30-40 minutes on an NVIDIA RTX 2080Ti.
Our model itself can generate images in real-time and takes about 0.01 seconds for one image on an NVIDIA RTX 2080Ti while one iteration of evaluating the mask connectivity takes about 0.00002 seconds on CPU.

\subsection{Quantitative Evaluation}
\vspace{-0.5em}
To the best of our knowledge there are no established quantitative evaluation metrics for few-shot character animation.
\cite{shaham2019singan} introduced the Single-Image FID (SIFID) score to evaluate single-image generative models.
However, \cite{hinz2020improved} report large variance in the SIFID scores and \cite{robb2020few} report that models in the few-shot setting overfit to the FID metric.
We use the Peak Signal-to-Noise Ratio (PSNR) and LPIPS \cite{zhang2018unreasonable} as metrics to evaluate our model.

To evaluate our model we design an $N$-fold cross-validation for a given character with $N$ images.
Given a character we train our model $N$ times on $N-1$ images were each image in the dataset is left out of training exactly once.
At test time we generate the left-out image based on its keypoint layout and calculate the PSNR and LPIPS between the generated and ground-truth image.
For each character we run the full $N$-fold cross-validation three times and report the average and standard deviation across the three runs.
Some examples are shown in \autoref{fig:reconstruction}.

\input{fig_layered_vs_baseline}

\autoref{tab:results:cross_val:baselines} shows the results of our model compared to the baselines.
Our model achieves the best LPIPS and PSNR for all characters.
We observe that the PSNR is not always predictive of the (perceptive) quality of the generated image.
In particular, SinGAN and ConSinGAN often generate images where the character exhibits disconnected body parts (e.g. the feet are not connected to the main body), but this is not represented in the PSNR, as feet and legs cover a relatively small area of the image.

\autoref{tab:results:ablation} shows ablation studies with our model, where we train the same model without any layering, and with layering but no adaptive scaling.
We can see that the layering approach improves the performance in all cases.
Finally, adding the adaptive scaling further improves the performance, albeit not as much as the keypoint layering.
While the difference in the quantitative evaluation may not seem large between our layered and baseline model, this is partly due to the quantitative metrics which evaluate the global image, while we show that the improvements are mainly in areas of overlapping keypoints, leading to clear qualitative improvements (e.g. see \autoref{fig:layered_vs_baseline}).

\vspace{-0.5em}
\subsection{Qualitative Evaluation}
\vspace{-0.5em}
\autoref{fig:our_model_examples} shows visualizations of our model's capabilities on several different characters.
All examples in this figure are trained on sprite sheets which contain 8 -- 12 examples of the given character.
Our model learns to generate realistic samples of four-legged animals, two-legged animals, and humanoid shapes.
Furthermore, we see that our model can handle the movement of keypoints that relate to relatively small character parts (e.g. individual feet) as well as keypoints that represent large body parts (e.g. head and torso).
Even when we move multiple keypoints, the resulting image is realistic, adheres to the novel keypoint layout, and leaves areas of the character that were not modified unchanged.

\input{fig_animation_layer_visualization}

\autoref{fig:teaser} and \autoref{fig:anims} show how our model can be used for character animation.
We use the poses from the training set as starting point and linearly interpolate the keypoints to generate the intermediate frames.
Our model produces smooth and realistic results. 
We note that these intermediate keypoint locations could also be derived from other data e.g.\ by extracting keypoints from driving videos \cite{aberman2018neural}.
For comparison, we also show the results of a recent general purpose frame-interpolation model DAIN \cite{bao2019depth}.

\myparagraph{Mask-based Keypoint Refinement}
When moving certain keypoints to a new location we sometimes observe that this leads to ``rips'' in the generated character since the keypoint is too far away from the main body and such a configuration does not occur in training data.
This can usually be fixed by moving the connected body part in the same direction as the original keypoint.
\autoref{fig:fix_kp} shows examples of rips in the generated character (first two rows) and introduced artifacts (third row).
The first column shows the original image and user specified modifications.
The second and third columns show the predicted mask and generated image.
The last column shows our model's final output after the respective connected keypoint was moved automatically.
We can see that by enforcing the mask connectivity constraint we can get more realistic results in all cases.

\myparagraph{Layered vs Non-layered Architecture}
\autoref{fig:layered_vs_baseline} shows qualitative comparisons between images generated by our architecture while using either our layered approach or a non-layered approach.
As discussed previously, our layered approach is beneficial when several keypoints overlap in the 2D space, e.g.\ if a hand passes in front of a body or if two body parts are very close to each other.
\autoref{fig:animation_layers} visualizes the features that our model learns for each keypoint layer.
As we can see the model learns to only model the relevant keypoints and their associated body parts for each layer.

\myparagraph{Human Reposing}
\autoref{fig:animation_real_person} shows how our model, trained on 10 images of the given character, compares to human reposing models which are pretrained on large datasets.
To make the comparison fairer we chose a relatively ``simple'' human character, i.e., with a standard body shape and tight clothing.
The first two models (PATN \cite{zhu2019progressive} and XinGAN \cite{tang2020xinggan}) are pretrained on the Market-1501 dataset \cite{zheng2015scalable} and are not capable of modeling this human who is slightly outside of the original training distribution.
We also evaluated these models pretrained on the DeepFashion dataset \cite{liuLQWTcvpr16DeepFashion} but the performance was worse.
The LWG \cite{liu2020liquid}, pretrained on the iPER dataset \cite{lwb2019}, performs a fine-tuning step on the same training images as we use before running the animation and is much better at modeling this person.
These results indicate that current human reposing models are not (yet) capable of generalizing to humans outside of their training distribution unless a fine-tuning step on the individual human is performed, hinting at the possibility of using our approach for modeling specific humans or use-cases.
We assume that future human reposing models will be able to model examples similar to the one chosen here and, therefore, suggest that our approach might be better suited to model highly individual persons (e.g., wearing unconventional clothes) or a person performing a very specific movement.

\input{fig_animation_real_person}
\input{fig_animation_discrete_location}
\input{fig_fix_kp_with_mask}

\myparagraph{Automatic Appearance Switching}
Our model does not only learn to associate discrete keypoints with given features, but also learns to associate different features with a keypoint based on its location relative to other keypoints. 
\autoref{fig:discrete_locations} shows several examples in which we can see that individual parts get  ``flipped'' as a function of the location of that keypoint with respect to the others. 
As in our previous examples, the features of unrelated keypoints are unaffected by this.

\myparagraph{Appearance-specific Keypoints}
\autoref{fig:discrete_keypoints} shows how our model is able to switch between discrete appearance states for given keypoints.
Each of the characters shows different visual features during training which we encode as different keypoint conditions.
At test time we can switch between these different states to combine novel poses with any of the discrete visual expressions.
Note that, again, our model learns to associate good features with the given keypoints, allowing us to model the poses independently of the discrete keypoint states.

\input{fig_discrete_keypoints}
\input{fig_animation_scale_dataset}

\myparagraph{Scaling to Larger Datasets}
While we show that our model performs well with only 8 -- 15 training images, we also evaluate our model on characters for which we have more training images ($\ge 50$).
\autoref{fig:animation_scale_dataset} shows how our model scales with larger datasets.
We see that more data is especially helpful when there are overlaps and occlusions.
While the models trained on only 5 or 15 images have difficulty modeling this (e.g.\ when the hand moves in front of the body), the models which are trained on more images perform much better.

\myparagraph{Limitations and Future Work}
Our model addresses the main challenge of correctly modeling (dis)occlusions based on limited training data.
However, our model still has no explicit understanding about any underlying 3D representation of the character and \autoref{fig:animation_scale_dataset} shows how modeling occlusions gets worse with fewer training images.
Extending the model to handle 3D characters is an interesting avenue for future work.
We hypothesize that this may benefit from additional information available in 3D models (such as camera pose).
Additionally, the concept of layering would have to be made more adaptive since different viewpoints would necessitate different layers of keypoints.

\vspace{-0.5em}
\section{Conclusion}
\vspace{-0.5em}
In this work we show how to train GANs on few examples (8 -- 15 images) of a given character for few-shot character reposing and animation.
The model is easy to use, requires no expert knowledge, and our layering approach produces realistic results for novel poses and occlusions.
We can perform character reposing in real-time through moving around keypoints and can animate character by interpolating keypoints.
In the future, this can be used with other approaches, e.g.\ by extracting keypoints from driving videos for character animation.
Through the use of a predicted foreground mask we can also automatically fix keypoint layouts that lead to unrealistic character poses.
Finally, we show that our model learns discrete state changes based on keypoint locations, associates keypoints and their layers with semantic body parts, and scales to larger datasets.

\section*{Acknowledgements}
\vspace{-0.5em}
\noindent The authors gratefully acknowledge partial support from the German Research Foundation DFG under project CML (TRR 169). We also thank Zuzana Studen\'a who produced some of the artwork used in this work.

{\small
\bibliographystyle{ieee_fullname}
\bibliography{egbib}
}

\end{document}

%% file: fig_teaser.tex
\newlength{\itemheight}
{
\centering
\renewcommand{\tabcolsep}{3pt}
\setlength{\itemheight}{2.275cm}
\begin{tabular}{ccccccc}
\animategraphics[autoplay,loop,height=\itemheight]{20}{figures/animations/Dog/CharacterGAN/gen_img_}{0000}{0110} &
\animategraphics[autoplay,loop,height=\itemheight]{10}{figures/animations/Dino/CharacterGAN/gen_img_}{0000}{0060} &
\animategraphics[autoplay,loop,height=\itemheight]{20}{figures/animations/Ostrich/CharacterGAN/gen_img_}{0000}{0090} &
\animategraphics[autoplay,loop,height=\itemheight]{20}{figures/animations/Man/CharacterGAN/gen_img_}{0000}{0080} &
\animategraphics[autoplay,loop,height=\itemheight]{10}{figures/animations/Ape/CharacterGAN/pm_gen_img_}{0000}{0050} &
\animategraphics[autoplay,loop,height=\itemheight]{20}{figures/animations/Fox/gen_img_}{0000}{0060} &
\animategraphics[autoplay,loop,height=\itemheight]{20}{figures/animations/Cow/CharacterGAN/gen_img_}{0000}{0080} \\
\end{tabular}
\renewcommand{\arraystretch}{0.5}
\renewcommand{\tabcolsep}{1pt}
\setlength{\itemheight}{1.5cm}
\begin{tabular}{ccccccccccc}
     \fbox{\includegraphics[height=\itemheight]{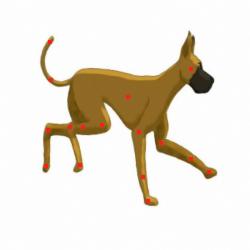}} &
     \includegraphics[height=\itemheight]{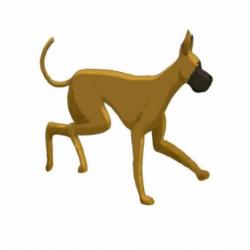} &
     \includegraphics[height=\itemheight]{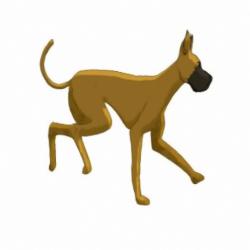} &
     \includegraphics[height=\itemheight]{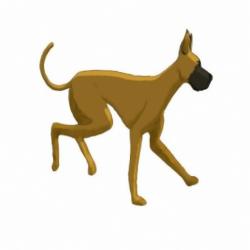} &
     \includegraphics[height=\itemheight]{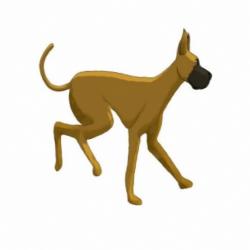} &
     \includegraphics[height=\itemheight]{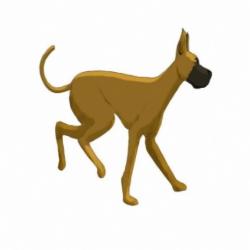} &
     \includegraphics[height=\itemheight]{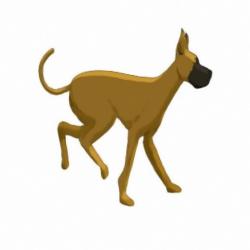} &
     \includegraphics[height=\itemheight]{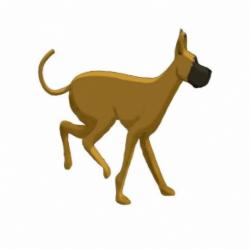} &
     \includegraphics[height=\itemheight]{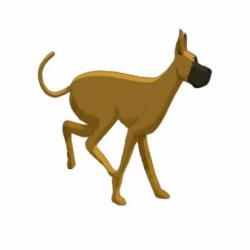} &
     \includegraphics[height=\itemheight]{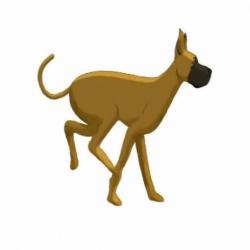} &
     \fbox{\includegraphics[height=\itemheight]{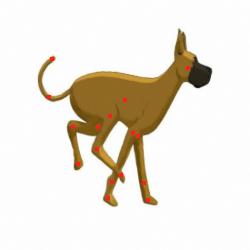}} \\
     
     \fbox{\includegraphics[height=\itemheight]{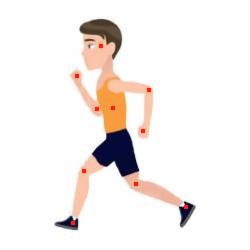}} &
     \includegraphics[height=\itemheight]{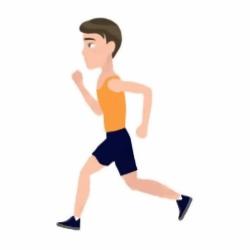} &
     \includegraphics[height=\itemheight]{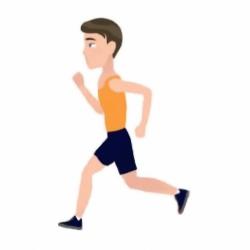} &
     \includegraphics[height=\itemheight]{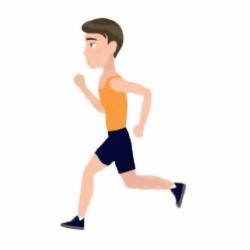} &
     \includegraphics[height=\itemheight]{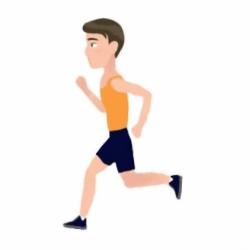} &
     \includegraphics[height=\itemheight]{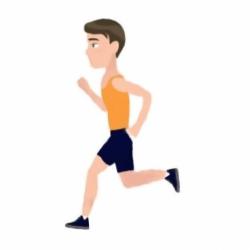} &
     \includegraphics[height=\itemheight]{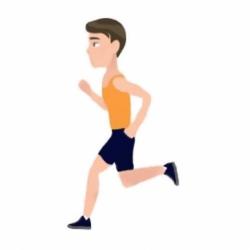} &
     \includegraphics[height=\itemheight]{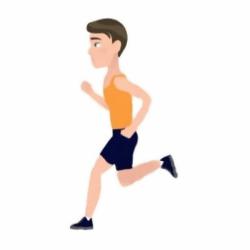} &
     \includegraphics[height=\itemheight]{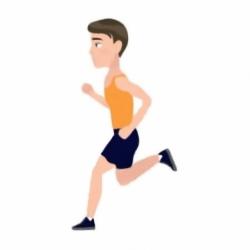} &
     \includegraphics[height=\itemheight]{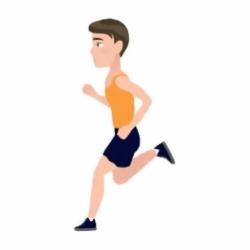} &
     \fbox{\includegraphics[height=\itemheight]{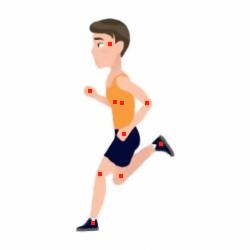}} \\
     Input & & & & & & & & & & Input \\
\end{tabular}
}

%% file: tab_baselines.tex
\begin{table*}
    \centering
    \renewcommand{\tabcolsep}{3pt}
    \small
    \begin{tabular}{l cc cc cc cc}
         \toprule
         Dataset & \multicolumn{2}{c}{SinGAN \cite{shaham2019singan}} & \multicolumn{2}{c}{ConSinGAN \cite{hinz2020improved}} & \multicolumn{2}{c}{DeepSIM \cite{vinker2020deep}} & \multicolumn{2}{c}{CharacterGAN} \\
         & PSNR $\uparrow$ & LPIPS $\downarrow$ & PSNR $\uparrow$ & LPIPS $\downarrow$ & PSNR $\uparrow$ & LPIPS $\downarrow$ & PSNR $\uparrow$ & LPIPS $\downarrow$ \\
         \midrule
         Watercolor Man & 18.03$\pm$0.09 & 0.149$\pm$0.0032 & 21.65$\pm$0.32 & 0.102$\pm$0.0151 & 21.92$\pm$0.03 & 0.066$\pm$0.0002 & \textbf{24.33$\pm$0.05} & \textbf{0.042$\pm$0.0002} \\
         Ape & 13.94$\pm$0.02 & 0.234$\pm$0.0098 & 18.09$\pm$0.08 & 0.132$\pm$0.0058 & 17.01$\pm$0.01 & 0.132$\pm$0.0009 & \textbf{18.94$\pm$0.04} & \textbf{0.089$\pm$0.0004} \\
         Sprite Man & 17.14$\pm$0.08 & 0.156$\pm$0.0063 & 23.19$\pm$0.17 & 0.087$\pm$0.0045 & 22.08$\pm$0.07 & 0.071$\pm$0.0001 & \textbf{25.23$\pm$0.02} & \textbf{0.038$\pm$0.0006} \\
         Dog & 15.01$\pm$0.28 & 0.195$\pm$0.0078 & 19.24$\pm$0.27 & 0.125$\pm$0.0089 & 20.08$\pm$0.07 & 0.087$\pm$0.0010 & \textbf{22.22$\pm$0.04} & \textbf{0.062$\pm$0.0006} \\
         Ostrich & 16.54$\pm$0.03 & 0.200$\pm$0.0023 & 23.30$\pm$0.18 & 0.103$\pm$0.0020 & 21.54$\pm$0.13 & 0.079$\pm$0.0006 & \textbf{23.80$\pm$0.09} & \textbf{0.063$\pm$0.0005} \\
         Cow & 13.58$\pm$0.03 & 0.220$\pm$0.0016 & 18.71$\pm$0.12 & 0.133$\pm$0.0037 & 17.65$\pm$0.03 & 0.115$\pm$0.0013 & \textbf{19.59$\pm$0.01} & \textbf{0.085$\pm$0.0005} \\
         \bottomrule
    \end{tabular}
    
    \vspace{-1em}
    \caption{Results of cross validation for the different models.}
    \label{tab:results:cross_val:baselines}
\end{table*}

%% file: tab_ablation.tex
\begin{table*}
    \centering
    \renewcommand{\tabcolsep}{5pt}
    \small
    \def\qualheight{1.045cm}
    \begin{tabular}{l cc cc}
         \toprule
          & \multicolumn{2}{c}{CharacterGAN} & \multicolumn{2}{c}{CharacterGAN} \\
         Dataset & \multicolumn{2}{c}{No Layer, No Scaling} & \multicolumn{2}{c}{Layer, No Scaling} \\
         & PSNR $\uparrow$ & LPIPS $\downarrow$ & PSNR $\uparrow$ & LPIPS $\downarrow$ \\
         \midrule
         Watercolor Man & 23.81 $\pm$ 0.15 & 0.049 $\pm$ 0.0050 & 24.29 $\pm$ 0.02 & 0.044 $\pm$ 0.0004  \\
         Ape & 18.11 $\pm$ 0.06 & 0.109 $\pm$ 0.0009 & 18.92 $\pm$ 0.03 & 0.092 $\pm$ 0.0003 \\
         Sprite Man & 24.63 $\pm$ 0.12 & 0.041 $\pm$ 0.0020 & 25.12 $\pm$ 0.01 & 0.039 $\pm$ 0.0004\\
         Dog & 21.71 $\pm$ 0.10 & 0.068 $\pm$ 0.0007 & 22.14 $\pm$ 0.05 & 0.064 $\pm$ 0.0005 \\
         Ostrich & 22.95 $\pm$ 0.04 & 0.068 $\pm$ 0.0007 & 22.97 $\pm$ 0.04 & 0.067 $\pm$ 0.0004 \\
         Cow & 18.95 $\pm$ 0.08 & 0.094 $\pm$ 0.0021 & 19.52 $\pm$ 0.01 & 0.086 $\pm$ 0.0009 \\
         \bottomrule
    \end{tabular}
    \begin{tabular}{ccc}
    \toprule
    Watercolor Man & Ape & Ostrich \\
    & & \\
    Sprite Man & Dog & Cow \\
    \midrule
    \includegraphics[height=\qualheight]{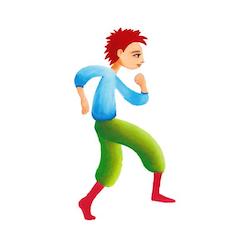}& 
    \includegraphics[height=\qualheight]{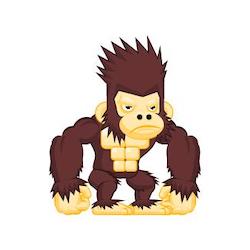}& 
    \includegraphics[height=\qualheight]{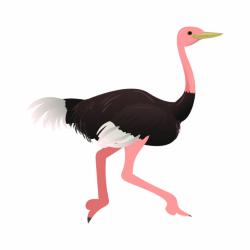} \\
    \includegraphics[height=\qualheight]{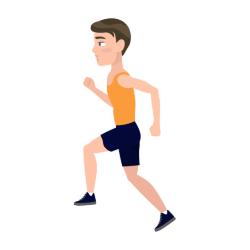}&
    \includegraphics[height=\qualheight]{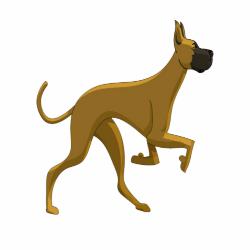}&
    \includegraphics[height=\qualheight]{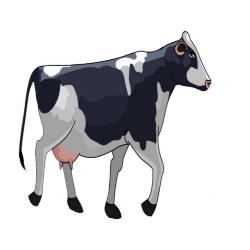} \\
    \bottomrule
    \end{tabular}
    
    \vspace{-1em}
    \caption{Ablation study: results of cross validation for different parts of our model.}
    \vspace{-1em}
    \label{tab:results:ablation}
\end{table*}

%% file: fig_our_model_examples.tex
\begin{figure*}[tb]
    \centering
    \renewcommand{\arraystretch}{0.6}
    \def\qualheight{1.9cm}
    \def\fsh{\small}
    
    \begin{tabular}{*{2}{c@{\hspace{0px}}}p{1px} *{2}{c@{\hspace{0px}}}p{1px} *{2}{c@{\hspace{0px}}}p{1px} *{2}{c@{\hspace{0px}}}}
    \fsh Original & \fsh Generated && \fsh Original & \fsh Generated && \fsh Original & \fsh Generated && \fsh Original & \fsh Generated \\
    
    \includegraphics[height=\qualheight]{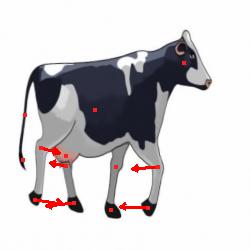}&
    \includegraphics[height=\qualheight]{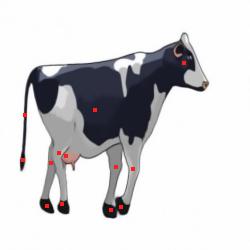}&&
    \includegraphics[height=\qualheight]{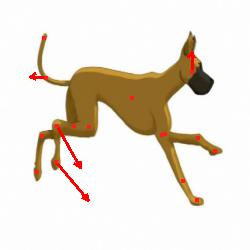}&
    \includegraphics[height=\qualheight]{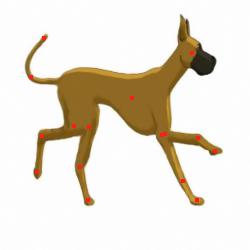}&&
    \includegraphics[height=\qualheight]{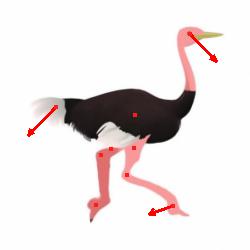}&
    \includegraphics[height=\qualheight]{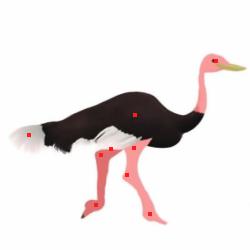}&&
    \includegraphics[height=\qualheight]{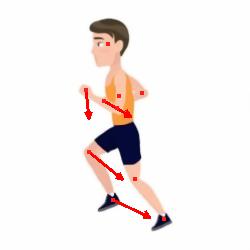}&
    \includegraphics[height=\qualheight]{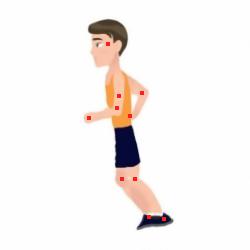}\\

    \end{tabular}

    \vspace{-0.5em}
    \caption{Examples from our model which was trained on only 8 -- 12 images for each of the characters. Odd columns show the original image and our intended modifications, even columns show the output of our model.}
    \vspace{-1em}
    \label{fig:our_model_examples}
\end{figure*}

%% file: fig_reconstruction.tex
\begin{figure}[tb]
    \centering
    \renewcommand{\arraystretch}{0.5}
    \renewcommand{\tabcolsep}{1pt}
    \def\qualheight{1.3cm}
    \def\fsh{\footnotesize}
    
    \begin{tabular}{*{6}{c@{\hspace{1px}}}}
    \fsh SinGAN & \fsh ConSinGAN & \fsh DeepSIM  & \multicolumn{2}{c}{\fsh CharacterGAN} & \fsh Ground \\
    & & & \tiny w/o layer & \tiny w/ layer & \fsh Truth \\
    
    \includegraphics[height=\qualheight]{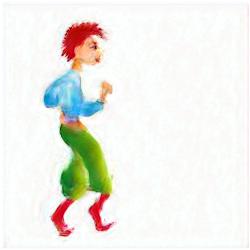}&
    \includegraphics[height=\qualheight]{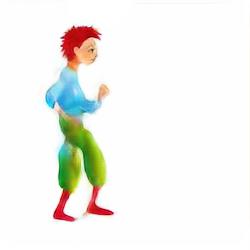}&
    \includegraphics[height=\qualheight]{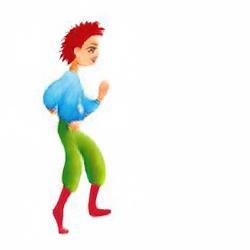}&
    \includegraphics[height=\qualheight]{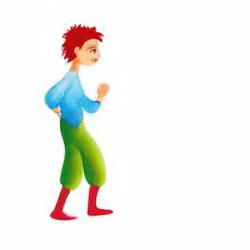}&
    \includegraphics[height=\qualheight]{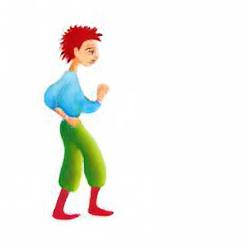}&
    \includegraphics[height=\qualheight]{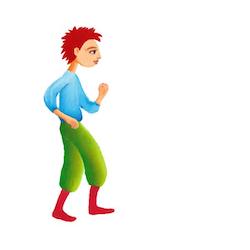}\\
    
    \includegraphics[height=\qualheight]{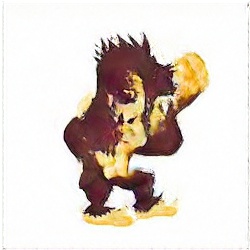}&
    \includegraphics[height=\qualheight]{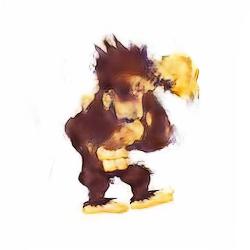}&
    \includegraphics[height=\qualheight]{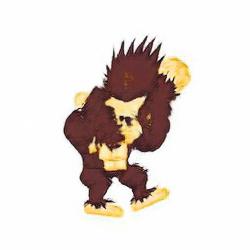}&
    \includegraphics[height=\qualheight]{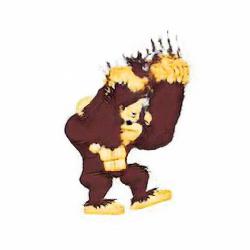}&
    \includegraphics[height=\qualheight]{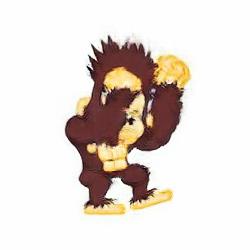}&
    \includegraphics[height=\qualheight]{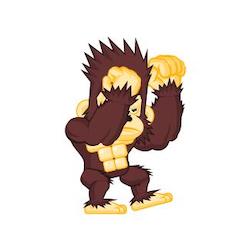}\\
    
    \includegraphics[height=\qualheight]{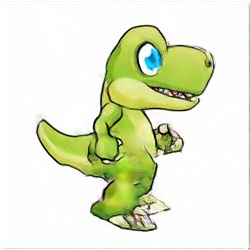}&
    \includegraphics[height=\qualheight]{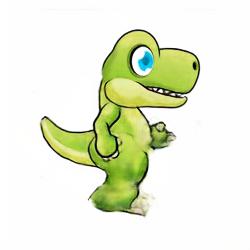}&
    \includegraphics[height=\qualheight]{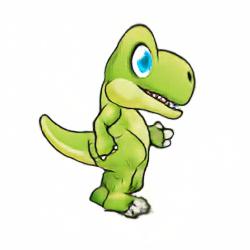}&
    \includegraphics[height=\qualheight]{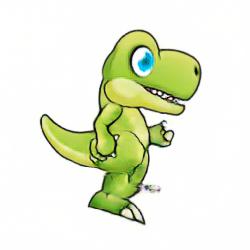}&
    \includegraphics[height=\qualheight]{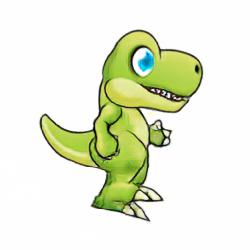}&
    \includegraphics[height=\qualheight]{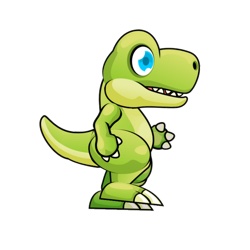}\\
    
    \end{tabular}
    
    \vspace{-0.5em}
    \caption{Qualitative examples of reconstructing held-out test images based on their keypoint locations.}
    \vspace{-1.5em}
    \label{fig:reconstruction}
\end{figure}

%% file: fig_animation.tex
\begin{figure*}[t]
\centering
\setlength{\itemheight}{1.74cm}
\renewcommand{\arraystretch}{0.6}
\renewcommand{\tabcolsep}{5pt}
\def\fsh{\small}
\begin{tabular}{ccccp{2px}cccc}
\animategraphics[autoplay,loop,height=\itemheight]{10}{figures/animations/Dog/DAIN/gen_img_}{00000}{00110} &
\animategraphics[autoplay,loop,height=\itemheight]{10}{figures/animations/Dog/DeepSIM/gen_img_}{0000}{0110} &
\animategraphics[autoplay,loop,height=\itemheight]{10}{figures/animations/Dog/CharacterGAN/gen_img_}{0000}{0110} &
\animategraphics[autoplay,loop,height=\itemheight]{10}{figures/animations/Dog/Real-Images/}{0000}{0110} &&
\animategraphics[autoplay,loop,height=\itemheight]{10}{figures/animations/Man/DAIN/gen_img_}{00000}{00080} &
\animategraphics[autoplay,loop,height=\itemheight]{10}{figures/animations/Man/DeepSIM/gen_img_}{0000}{0080} &
\animategraphics[autoplay,loop,height=\itemheight]{10}{figures/animations/Man/CharacterGAN/gen_img_}{0000}{0080} &
\animategraphics[autoplay,loop,height=\itemheight]{10}{figures/animations/Man/Real-Images/}{0000}{0080} \\

\animategraphics[autoplay,loop,height=\itemheight]{10}{figures/animations/Watercolor_Man/DAIN/gen_img_}{0000}{0100} &
\animategraphics[autoplay,loop,height=\itemheight]{10}{figures/animations/Watercolor_Man/DeepSIM/pm_gen_img_}{0000}{0100} &
\animategraphics[autoplay,loop,height=\itemheight]{10}{figures/animations/Watercolor_Man/CharacterGAN-Layered-15-IMGs-PM/gen_img_}{0000}{0100} &
\animategraphics[autoplay,loop,height=\itemheight]{10}{figures/animations/Watercolor_Man/Real-Images/}{0000}{0100} &&
\animategraphics[autoplay,loop,height=\itemheight]{10}{figures/animations/Cow/DAIN/gen_img_}{0000}{0080} &
\animategraphics[autoplay,loop,height=\itemheight]{10}{figures/animations/Cow/DeepSIM/gen_img_}{0000}{0080} &
\animategraphics[autoplay,loop,height=\itemheight]{10}{figures/animations/Cow/CharacterGAN/gen_img_}{0000}{0080} &
\animategraphics[autoplay,loop,height=\itemheight]{10}{figures/animations/Cow/Real-Images/}{0000}{0080} \\

\animategraphics[autoplay,loop,height=\itemheight]{7}{figures/animations/Dino/DAIN/gen_img_}{0000}{0070} &
\animategraphics[autoplay,loop,height=\itemheight]{7}{figures/animations/Dino/DeepSIM/gen_img_}{0000}{0070} &
\animategraphics[autoplay,loop,height=\itemheight]{7}{figures/animations/Dino/CharacterGAN/gen_img_}{0000}{0070} &
\animategraphics[autoplay,loop,height=\itemheight]{7}{figures/animations/Dino/Real-Images/}{0000}{0070} &&
\animategraphics[autoplay,loop,height=\itemheight]{7}{figures/animations/Ape/DAIN/gen_img_}{0000}{0050} &
\animategraphics[autoplay,loop,height=\itemheight]{7}{figures/animations/Ape/DeepSIM/pm_gen_img_}{0000}{0050} &
\animategraphics[autoplay,loop,height=\itemheight]{7}{figures/animations/Ape/CharacterGAN/pm_gen_img_}{0000}{0050} &
\animategraphics[autoplay,loop,height=\itemheight]{7}{figures/animations/Ape/Real-Images/}{0000}{0050} \\

\fsh DAIN & \fsh DeepSIM & \fsh Ours & \fsh Input && \fsh DAIN & \fsh DeepSIM & \fsh Ours & \fsh Input\\
\end{tabular}
    
\vspace{-0.5em}
\caption{Comparison of our approach to frame interpolation \cite{bao2019depth} and DeepSIM \cite{vinker2020deep} using the displayed input frames (\animation).}
\vspace{-1em}
\label{fig:anims}
\end{figure*}

%% file: fig_layered_vs_baseline.tex
\begin{figure}[tb]
    \centering
    \renewcommand{\arraystretch}{0.6}
    \def\qualheight{1.8cm}
    \def\fsh{\small}
    
    \begin{tabular}{*{2}c@{\hspace{0px}}p{0px} *{2}{c@{\hspace{0px}}}}
    \fsh No Layer & \fsh Layer && \fsh No Layer & \fsh Layer \\
    
    \includegraphics[height=\qualheight]{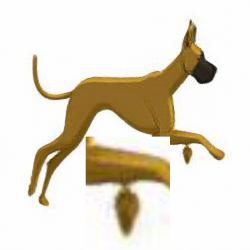}&
    \includegraphics[height=\qualheight]{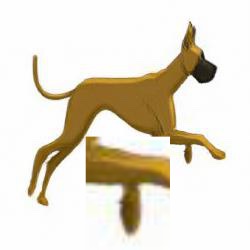}&&
    \includegraphics[height=\qualheight]{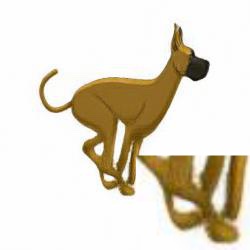}&
    \includegraphics[height=\qualheight]{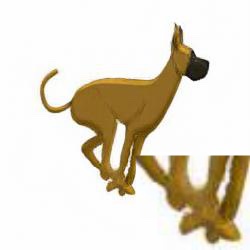}\\
    
    \includegraphics[height=\qualheight]{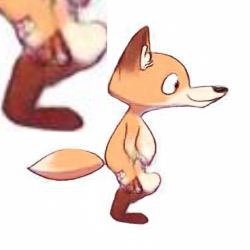}&
    \includegraphics[height=\qualheight]{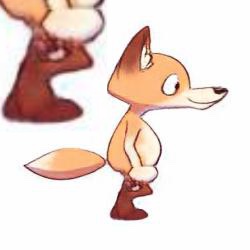}&&
    \includegraphics[height=\qualheight]{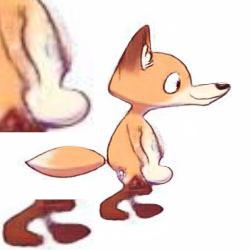}&
    \includegraphics[height=\qualheight]{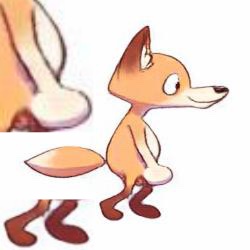}\\
    
    \includegraphics[height=\qualheight]{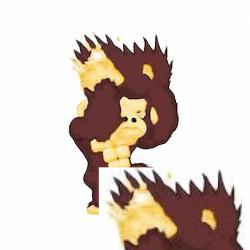}&
    \includegraphics[height=\qualheight]{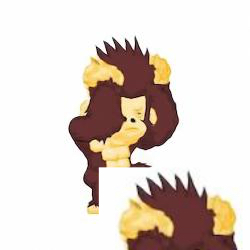}&&
    \includegraphics[height=\qualheight]{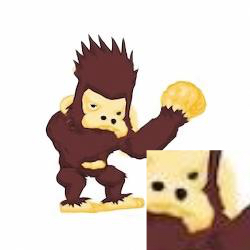}&
    \includegraphics[height=\qualheight]{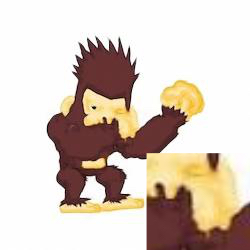}\\
    
    \includegraphics[height=\qualheight]{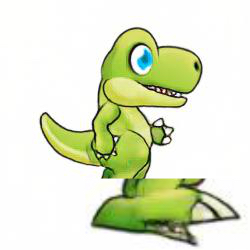}&
    \includegraphics[height=\qualheight]{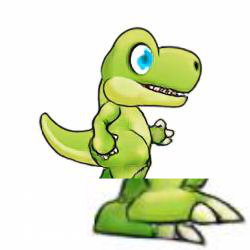}&&
    \includegraphics[height=\qualheight]{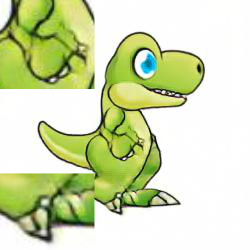}&
    \includegraphics[height=\qualheight]{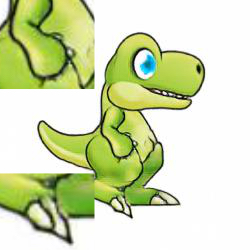}\\
    \end{tabular}

    \vspace{-0.5em}
    \caption{Comparison of layered vs non-layered at occlusions/overlaps.}
    \vspace{-1em}
    \label{fig:layered_vs_baseline}
\end{figure}

%% file: fig_animation_layer_visualization.tex
\begin{figure}[t]
\centering
\renewcommand{\arraystretch}{0.6}
\setlength{\itemheight}{1.4cm}
\begin{tabular}{*{1}{c@{\hspace{5px}}}p{10px} *{1}{c@{\hspace{5px}}}p{10px} *{1}{c@{\hspace{5px}}}}
\animategraphics[autoplay,loop,height=\itemheight]{10}{figures/animation_layers/Dog/Layer_0/gen_img_}{0000}{0110} &&
\animategraphics[autoplay,loop,height=\itemheight]{10}{figures/animation_layers/Dog/Layer_1/gen_img_}{0000}{0110} &&
\animategraphics[autoplay,loop,height=\itemheight]{10}{figures/animation_layers/Dog/Layer_2/gen_img_}{0000}{0110}  \\

\animategraphics[autoplay,loop,height=\itemheight]{10}{figures/animation_layers/Ostrich/Layer_0/gen_img_}{0000}{0080} &&
\animategraphics[autoplay,loop,height=\itemheight]{10}{figures/animation_layers/Ostrich/Layer_1/gen_img_}{0000}{0080} &&
\animategraphics[autoplay,loop,height=\itemheight]{10}{figures/animation_layers/Ostrich/Layer_2/gen_img_}{0000}{0080}  \\

\animategraphics[autoplay,loop,height=\itemheight]{10}{figures/animation_layers/Man/Layer_0/gen_img_}{0000}{0080} &&
\animategraphics[autoplay,loop,height=\itemheight]{10}{figures/animation_layers/Man/Layer_1/gen_img_}{0000}{0080} &&
\animategraphics[autoplay,loop,height=\itemheight]{10}{figures/animation_layers/Man/Layer_2/gen_img_}{0000}{0080}  \\

\end{tabular}
    
\vspace{-0.5em}
\caption{Visualization of what different layers learn (\animation).}
\vspace{-1.5em}
\label{fig:animation_layers}
\end{figure}

%% file: fig_animation_real_person.tex
\begin{figure}[t]
\centering
\renewcommand{\arraystretch}{0.6}
\setlength{\itemheight}{1.4cm}
\def\fsh{\small}

\begin{tabular}{*{1}{c@{\hspace{1px}}}p{1px} *{1}{c@{\hspace{1px}}}p{1px} *{1}{c@{\hspace{1px}}}p{1px} *{1}{c@{\hspace{1px}}}p{1px} *{1}{c@{\hspace{1px}}}}
\fsh PATN && \fsh XinGAN &&  \fsh LWG && \fsh Character && \fsh Original \\
\fsh \cite{zhu2019progressive} && \fsh \cite{tang2020xinggan} &&  \fsh \cite{liu2020liquid} && \fsh GAN &&  \\
\animategraphics[autoplay,loop,height=\itemheight]{9}{figures/animations_real_person/Pose-Transfer/gen_img_}{0000}{0089} &&
\animategraphics[autoplay,loop,height=\itemheight]{9}{figures/animations_real_person/XinGAN/gen_img_}{0000}{0089} &&
\animategraphics[autoplay,loop,height=\itemheight]{9}{figures/animations_real_person/iPERCore/gen_img_}{0000}{0089} &&
\animategraphics[autoplay,loop,height=\itemheight]{9}{figures/animations_real_person/CharacterGAN-Layered/gen_img_}{0000}{0089} &&
\animategraphics[autoplay,loop,height=\itemheight]{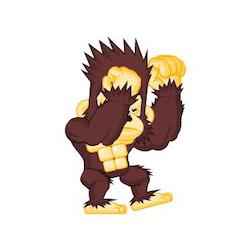}{figures/animations_real_person/Original/original_}{0000}{0009} \\
\end{tabular}
    
\vspace{-0.5em}
\caption{Animation of a real person (\animation).}
\vspace{-0.5em}
\label{fig:animation_real_person}
\end{figure}

%% file: fig_animation_discrete_location.tex
\begin{figure}[t]
\centering
\renewcommand{\arraystretch}{0.6}
\setlength{\itemheight}{1.4cm}
\begin{tabular}{*{1}{c@{\hspace{5px}}} *{1}{c@{\hspace{5px}}} *{1}{c@{\hspace{5px}}} *{1}{c@{\hspace{5px}}}*{1}{c@{\hspace{5px}}}}
\animategraphics[autoplay,loop,height=\itemheight]{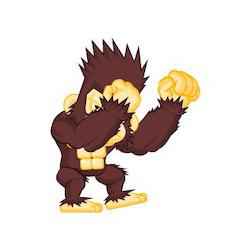}{figures/animations_discrete_locations/watercolor_guy_1/gen_img_}{0}{10} &
\animategraphics[autoplay,loop,height=\itemheight]{2}{figures/animations_discrete_locations/man/gen_img_}{0}{10} &
\animategraphics[autoplay,loop,height=\itemheight]{2}{figures/animations_discrete_locations/cow/gen_img_}{0}{10} &
\animategraphics[autoplay,loop,height=\itemheight]{2}{figures/animations_discrete_locations/strauss/gen_img_}{0}{10} &
\animategraphics[autoplay,loop,height=\itemheight]{2}{figures/animations_discrete_locations/maddy/gen_img_}{0}{10} \\
\end{tabular}
    
\vspace{-0.5em}
\caption{Discrete appearance changes (\animation).}
\vspace{-0.5em}
\label{fig:discrete_locations}
\end{figure}

%% file: fig_fix_kp_with_mask.tex
\begin{figure}[tb]
    \centering
    \renewcommand{\arraystretch}{0.6}
    \def\qualheight{1.9cm}
    \def\fsh{\small}
    
    \begin{tabular}{*{4}{c@{\hspace{0px}}}}
    \fsh Original & \fsh Predicted Mask & \fsh Before Fix & \fsh Fixed \\
    
    \includegraphics[height=\qualheight]{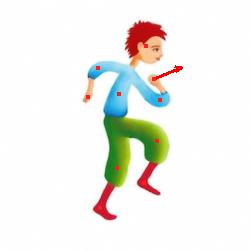}&
    \includegraphics[height=\qualheight]{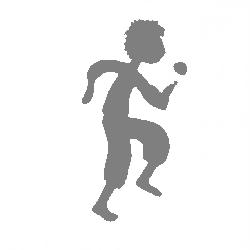}&
    \includegraphics[height=\qualheight]{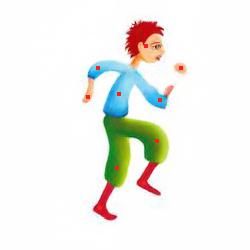}&
    \includegraphics[height=\qualheight]{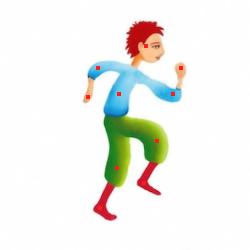}\\
    
    \includegraphics[height=\qualheight]{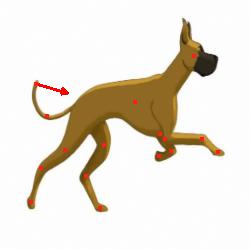}&
    \includegraphics[height=\qualheight]{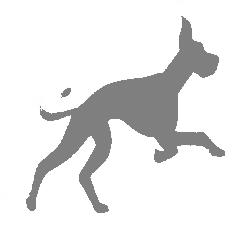}&
    \includegraphics[height=\qualheight]{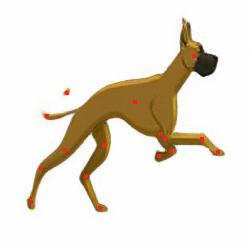}&
    \includegraphics[height=\qualheight]{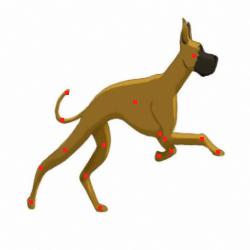}\\
    
    \includegraphics[height=\qualheight]{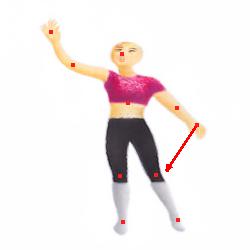}&
    \includegraphics[height=\qualheight]{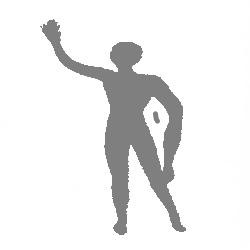}&
    \includegraphics[height=\qualheight]{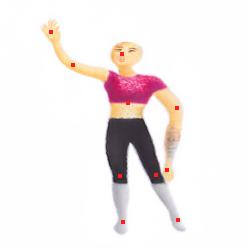}&
    \includegraphics[height=\qualheight]{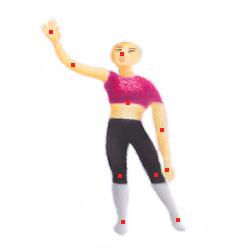}\\
    
    \end{tabular}

    \vspace{-0.5em}
    \caption{Enforcing mask connectivity at test time results in more realistic images.}
    \vspace{-1.0em}
    \label{fig:fix_kp}
\end{figure}

%% file: fig_discrete_keypoints.tex
\begin{figure}[tb]
    \centering
    \renewcommand{\arraystretch}{0.6}
    \def\qualheight{1.5cm}
    \def\fsh{\small}
    
    \begin{tabular}{*{2}{c@{\hspace{10px}}}p{10px} *{2}{c@{\hspace{10px}}}}
    \fsh Original & \fsh Generated && \fsh Original & \fsh Generated \\
    
    \includegraphics[height=\qualheight]{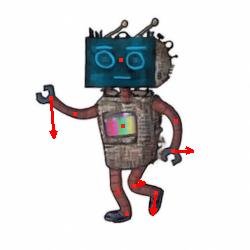}&
    \includegraphics[height=\qualheight]{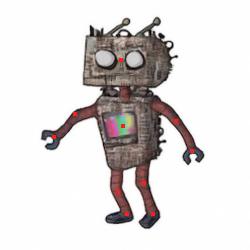}&&
    \includegraphics[height=\qualheight]{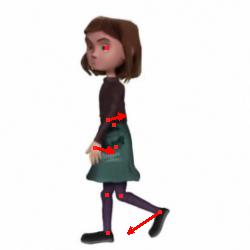}&
    \includegraphics[height=\qualheight]{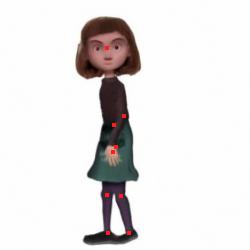}\\
    
    \end{tabular}

    \vspace{-0.5em}
    \caption{Discrete appearance change based on keypoint selection. In this example, the user not only moves keypoints to generate a new pose, but switches the IDs of keypoints to change expression, such as the color and rotation of the characters' faces.}
    \vspace{-1em}
    \label{fig:discrete_keypoints}
\end{figure}

%% file: fig_animation_scale_dataset.tex
\begin{figure}[t]
\centering
\renewcommand{\tabcolsep}{1pt}
\def\itemheight{1.35cm}
\def\fsh{\small}
\begin{tabular}{*{3}{c@{\hspace{0px}}}p{1px} *{3}{c@{\hspace{0px}}}}
\multicolumn{7}{c}{\fsh Number of Training Images} \\
\fsh 5 & \fsh 15 & \fsh 89&& \fsh 5 & \fsh 15 & \fsh 85 \\
\animategraphics[autoplay,loop,height=\itemheight]{15}{figures/animations/Watercolor_Man/CharacterGAN-Layered-5-IMGs-PM/gen_img_}{0030}{0100} &
\animategraphics[autoplay,loop,height=\itemheight]{15}{figures/animations/Watercolor_Man/CharacterGAN-Layered-15-IMGs-PM/gen_img_}{0030}{0100} &
\animategraphics[autoplay,loop,height=\itemheight]{15}{figures/animations/Watercolor_Man/CharacterGAN-Layered-89-IMGs-PM/gen_img_}{0030}{0100}&&
\animategraphics[autoplay,loop,height=\itemheight]{20}{figures/animations/Watercolor_Lady/CharacterGAN-Layer-5-IMGs/gen_img_}{0000}{0090} &
\animategraphics[autoplay,loop,height=\itemheight]{20}{figures/animations/Watercolor_Lady/CharacterGAN-Layer-15-IMGs/gen_img_}{0000}{0090} &
\animategraphics[autoplay,loop,height=\itemheight]{20}{figures/animations/Watercolor_Lady/CharacterGAN-Layer-85-IMGs/gen_img_}{0000}{0090} \\ \end{tabular}
    
\vspace{-0.5em}
\caption{Here we show how performance improves with more training images (\animation ).}
\vspace{-1em}
\label{fig:animation_scale_dataset}
\end{figure}